# Representations of language in a model of visually grounded speech signal

**Grzegorz Chrupała**
Tilburg University
g.chrupala@uvt.nl

**Lieke Gelderloos**
Tilburg University
l.j.gelderloos@uvt.nl

**Afra Alishahi**
Tilburg University
a.alishahi@uvt.nl

## Abstract

We present a visually grounded model of speech perception which projects spoken utterances and images to a joint semantic space. We use a multi-layer recurrent highway network to model the temporal nature of spoken speech, and show that it learns to extract both form and meaning-based linguistic knowledge from the input signal. We carry out an in-depth analysis of the representations used by different components of the trained model and show that encoding of semantic aspects tends to become richer as we go up the hierarchy of layers, whereas encoding of form-related aspects of the language input tends to initially increase and then plateau or decrease.

## 1 Introduction

Speech recognition is one of the success stories of language technology. It works remarkably well in a range of practical settings. However, this success relies on the use of very heavy supervision where the machine is fed thousands of hours of painstakingly transcribed audio speech signal. Humans are able to learn to recognize and understand speech from notably weaker and noisier supervision: they manage to learn to extract structure and meaning from speech by simply being exposed to utterances situated and grounded in their daily sensory experience. Modeling and emulating this remarkable skill has been the goal of numerous studies; however in the overwhelming majority of cases researchers used severely simplified settings where either the language input or the extralinguistic sensory input, or both, are small scale and symbolically represented. Section 2 provides a brief overview of this research.

More recently several lines of work have moved towards more realistic inputs while modeling or emulating language acquisition in a grounded setting. Gelderloos and Chrupała (2016) use the image captioning dataset MS COCO (Lin et al., 2014) to mimic the setting of grounded language learning: the sensory input consists of images of natural scenes, while the language input are phonetically transcribed descriptions of these scenes. The use of such moderately large and low-level data allows the authors to train a multi-layer recurrent neural network model, and to explore the nature and localization of the emerging hierarchy of linguistic representations learned in the process. Furthermore, in a series of recent studies Harwath and Glass (2015); Harwath et al. (2016); Harwath and Glass (2017) use image captioning datasets to model learning to understand spoken language from visual context with convolutional neural network models. Finally, there is a small but growing body of work dedicated to elucidating the nature of representations learned by neural networks from language data (see Section 2.2 for a brief overview). In the current work we build on these three strands of research and contribute the following advances:

- We use a multi-layer gated recurrent neural network to properly model the temporal nature of speech signal and substantially improve performance compared to the convolutional architecture from Harwath and Glass (2015);
- We carry out an in-depth analysis of the representations used by different components of the trained model and correlate them to representations learned by a text-based model and to human patterns of judgment on linguistic stimuli. This analysis is especially novel for a model with speech signal as input.

The general pattern of findings in our analysis is as follows: The model learns to extract from the acoustic input both form-related and semantics-related information, and encodes it in the activations of the hidden layers. Encoding of semantic aspects tends to become richer as we go up the hierarchy of layers. Meanwhile, encoding of form-related aspects of the language input, such as utterance length or the presence of specific words, tends to initially increase and then decay.

We release the code for our models and analyses as open source, available at https://github.com/gchrupala/visually-grounded-speech. We also release a dataset of synthetically spoken image captions based on MS COCO, available at https://doi.org/10.5281/zenodo.400926.

## 2 Related work

Children learn to recognize and assign meaning to words from continuous perceptual data in extremely noisy context. While there have been many computational studies of human word meaning acquisition, they typically make strong simplifying assumptions about the nature of the input. Often language input is given in the form of word symbols, and the context consists of a set of symbols representing possible referents (e.g. Siskind, 1996; Frank et al., 2007; Fazly et al., 2010). In contrast, several studies presented models that learn from sensory rather than symbolic input, which is rich with regards to the signal itself, but very limited in scale and variation (e.g. Roy and Pentland, 2002; Yu and Ballard, 2004; Lazaridou et al., 2016).

### 2.1 Multimodal language acquisition

Chrupała et al. (2015) introduce a model that learns to predict the visual context from image captions. The model is trained on image-caption pairs from MSCOCO (Lin et al., 2014), capturing both rich visual input as well as larger scale input, but the language input still consists of word symbols. Gelderloos and Chrupała (2016) propose a similar architecture that instead takes phoneme-level transcriptions as language input, thereby incorporating the word segmentation problem into the learning task. In this work, we introduce an architecture that learns from continuous speech and images directly.

This work is related to research on visual grounding of language. The field is large and growing, with most work dedicated to the grounding of written text, particularly in image captioning tasks (see Bernardi et al. (2016) for an overview). However, learning to ground language to visual information is also interesting from an automatic speech recognition point of view. Potentially, ASR systems could be trained from naturally co-occurring visual context information, without the need for extensive manual annotation – a particularly promising prospect for speech recognition in low-resource languages. There have been several attempts along these lines. Synnaeve et al. (2014) present a method of learning to recognize spoken words in isolation from co-occurrence with image fragments. Harwath and Glass (2015) present a model that learns to map pre-segmented spoken words in sequence to aspects of the visual context, while in Harwath and Glass (2017) the model also learns to recognize words in the unsegmented signal.

Most closely related to our work is that of Harwath et al. (2016), as it presents an architecture that learns to project images and unsegmented spoken captions to the same embedding space. The sentence representation is obtained by feeding the spectrogram to a convolutional network. The architecture is trained on crowd-sourced spoken captions for images from the Places dataset (Zhou et al., 2014), and evaluated on image search and caption retrieval. Unfortunately this dataset is not currently available and we were thus unable to directly compare the performance of our model to Harwath et al. (2016). We do compare to Harwath and Glass (2015) which was tested on a public dataset. We make different architectural choices, as our models are based on recurrent highway networks (Zilly et al., 2016). As in human cognition, speech is processed incrementally. This also allows our architecture to integrate information sequentially from speech of arbitrary duration.

### 2.2 Analysis of neural representations

While analysis of neural methods in NLP is often limited to evaluation of the performance on the training task, recently methods have been introduced to *peek inside the black box* and explore what it is that enables the model to perform the task. One approach is to look at the contribution of specific parts of the input, or specific units in the model, to final representations or decisions. Kádár et al. (2016) propose *omission scores*, a method to estimate the contribution of input tokens to the fi-nal representation by removing them from the input and comparing the resulting representations to the ones generated by the original input. In a similar approach, Li et al. (2016) study the contribution of individual input tokens as well as hidden units and word embedding dimensions by erasing them from the representation and analyzing how this affects the model.

Miao et al. (2016) and Tang et al. (2016) use visualization techniques for fine-grained analysis of GRU and LSTM models for ASR. Visualization of input and forget gate states allows Miao et al. (2016) to make informed adaptations to gated recurrent architectures, resulting in more efficiently trainable models. Tang et al. (2016) visualize qualitative differences between LSTM- and GRU-based architectures, regarding the encoding of information, as well as how it is processed through time.

We specifically study linguistic properties of the information encoded in the trained model. Adi et al. (2016) introduce prediction tasks to analyze information encoded in sentence embeddings about word order, sentence length, and the presence of individual words. We use related techniques to explore encoding of aspects of form and meaning within components of our stacked architecture.

## 3 Models

We use a multi-layer, gated recurrent neural network (RHN) to model the temporal nature of speech signal. Recurrent neural networks are designed for modeling sequential data, and gated variants (GRUs, LSTMs) are widely used with speech and text in both cognitive modeling and engineering contexts. RHNs are a simple generalization of GRU networks such that the transform between time points can consist of several steps.

Our multimodal model projects spoken utterances and images to a joint semantic space. The idea of projecting different modalities to a shared semantic space via a pair of encoders has been used in work on language and vision (among them Vendrov et al. (2015)). The core idea is to encourage inputs representing the same meaning in different modalities to end up nearby, while maintaining a distance from unrelated inputs.

The model consists of two parts: an utterance encoder, and an image encoder. The utterance encoder starts from MFCC speech features, while the image encoder starts from features extracted with a VGG-16 pre-trained on ImageNet. Our loss function attempts to make the cosine distance between encodings of matching utterances and images less than the distance between encodings of mismatching utterance/image pairs, by a margin:

$$\sum_{u,i} \left( \sum_{u'} \max[0, \alpha + d(u,i) - d(u',i)] + \sum_{i'} \max[0, \alpha + d(u,i) - d(u,i')] \right) \quad (1)$$

where $d(u,i)$ is the cosine distance between the encoded utterance $u$ and encoded image $i$. Here $(u,i)$ is the matching utterance-image pair, $u'$ ranges over utterances not describing $i$ and $i'$ ranges over images not described by $u$.

The image encoder $\text{enc}_i$ is a simple linear projection, followed by normalization to unit L2 norm:

$$\text{enc}_i(\mathbf{i}) = \text{unit}(\mathbf{A}\mathbf{i} + b) \quad (2)$$

where $\text{unit}(x) = \frac{x}{(x^T x)^{0.5}}$ and with $(\mathbf{A}, b)$ as learned parameters. The utterance encoder $\text{enc}_u$ consists of a 1-dimensional convolutional layer of length $s$, size $d$ and stride $z$, whose output feeds into a Recurrent Highway Network with $k$ layers and $L$ microsteps, whose output in turn goes through an attention-like lookback operator, and finally L2 normalization:

$$\text{enc}_u(\mathbf{u}) = \text{unit}(\text{Attn}(\text{RHN}_{k,L}(\text{Conv}_{s,d,z}(\mathbf{u})))) \quad (3)$$

The main function of the convolutional layer $\text{Conv}_{s,d,z}$ is to subsample the input along the temporal dimension. We use a 1-dimensional convolution with full border mode padding. The attention operator simply computes a weighted sum of the RHN activation at all timesteps:

$$\text{Attn}(\mathbf{x}) = \sum_t \alpha_t \mathbf{x}_t \quad (4)$$

where the weights $\alpha_t$ are determined by learned parameters $\mathbf{U}$ and $\mathbf{W}$, and passed through the timewise softmax function:

$$\alpha_t = \frac{\exp(\mathbf{U}\tanh(\mathbf{W}\mathbf{x}_t))}{\sum_{t'} \exp(\mathbf{U}\tanh(\mathbf{W}\mathbf{x}_{t'}))} \quad (5)$$

The main component of the utterance encoder is a recurrent network, specifically a Recurrent Highway Network (Zilly et al., 2016). The idea behind

RHN is to increase the depth of the transform between timesteps, or the recurrence depth. Otherwise they are a type of gated recurrent networks. The transition from timestep $t-1$ to $t$ is then defined as:

$$\text{rhn}(\mathbf{x}_t, \mathbf{s}_{t-1}^{(L)}) = \mathbf{s}_t^{(L)} \qquad (6)$$

where $\mathbf{x}_t$ stands for input at time $t$, and $\mathbf{s}_t^{(l)}$ denotes the state at time $t$ at recurrence layer $l$, with $L$ being the top layer of recurrence. Furthermore,

$$\mathbf{s}_t^{(l)} = \mathbf{h}_t^{(l)} \odot \mathbf{t}_t^{(l)} + \mathbf{s}_t^{(l-1)} \odot \left(\mathbf{1} - \mathbf{t}_t^{(l)}\right) \qquad (7)$$

where $\odot$ is elementwise multiplication, and

$$\mathbf{h}_t^{(l)} = \tanh\left(\mathbb{I}[l=1]\mathbf{W}_H \mathbf{x}_t + \mathbf{U}_{H_l}\mathbf{s}_t^{(l-1)}\right) \qquad (8)$$

$$\mathbf{t}_t^{(l)} = \sigma\left(\mathbb{I}[l=1]\mathbf{W}_T \mathbf{x}_t + \mathbf{U}_{T_l}\mathbf{s}^{(l-1)}\right) \qquad (9)$$

Here $\mathbb{I}$ is the indicator function: input is only included in the computation for the first layer of recurrence $l = 1$. By applying the rhn function repeatedly, an RHN layer maps a sequence of inputs to a sequence of states:

$$\begin{aligned}&\text{RHN}(\mathbf{X}, \mathbf{s}_0) \\ &= \text{rhn}(\mathbf{x}_n, \ldots, \text{rhn}(\mathbf{x}_2, \text{rhn}(\mathbf{x}_1, \mathbf{s}_0^{(L)})))\end{aligned} \qquad (10)$$

Two or more RHN layers can be composed into a stack:

$$\text{RHN}_2(\text{RHN}_1(\mathbf{X}, \mathbf{s_1}_0^{(L)}), \mathbf{s_2}_0^{(L)}), \qquad (11)$$

where $\mathbf{s_n}_t^{(l)}$ stands for the state vector of layer $n$ of the stack, at layer $l$ of recurrence, at time $t$. In our version of the Stacked RHN architecture we use *residualized* layers:

$$\text{RHN}_{\text{res}}(\mathbf{X}, \mathbf{s}_0) = \text{RHN}(\mathbf{X}, \mathbf{s}_0) + \mathbf{X} \qquad (12)$$

This formulation tends to ease optimization in multi-layer models (cf. He et al., 2015; Oord et al., 2016).

In addition to the speech model described above, we also define a comparable text model. As it takes a sequence of words as input, we replace the convolutional layer with a word embedding lookup table. We found the text model did not benefit from the use of the attention mechanism, and thus the sentence embedding is simply the L2-normalized activation vector of the topmost layer, at the last timestep.

## 4 Experiments

Our main goal is to analyze the emerging representations from different components of the model and to examine the linguistic knowledge they encode. For this purpose, we employ a number of tasks that cover the spectrum from fully form-based to fully semantic.

In Section 4.2 we assess the effectiveness of our architecture by evaluating it on the task of ranking images given an utterance. Sections 4.3 to 4.6 present our analyses. In Sections 4.3 and 4.4 we define auxiliary tasks to investigate to what extent the network encodes information about the surface form of an utterance from the speech input. In Section 4.5 and 4.6 we focus on where semantic information is encoded in the model. In the analyses, we use the following features:

**Utterance embeddings:** the weighted sum of the unit activations on the last layer, as calculated by Equation (3).

**Average unit activations:** hidden layer activations averaged over time and L2-normalized for each hidden layer.

**Average input vectors:** the MFCC vectors averaged over time. We use this feature to examine how much information can be extracted from the input signal only.

### 4.1 Data

For the experiments reported in the remainder of the paper we use two datasets of images with spoken captions.

#### 4.1.1 Flickr8K

The Flickr8k Audio Caption Corpus was constructed by having crowdsource workers read aloud the captions in the original Flickr8K corpus (Hodosh et al., 2013). For details of the data collection procedure refer to Harwath and Glass (2015). The datasets consist of 8,000 images, each image with five descriptions. One thousand images are held out for validation, and another one thousand for the final test set. We use the splits provided by (Karpathy and Fei-Fei, 2015). The image features come from the final fully connect layer of VGG-16 (Simonyan and Zisserman, 2014) pre-trained on Imagenet (Russakovsky et al., 2014).

We generate the input signal as follows: we extract 12-dimensional mel-frequency cepstral coefficients (MFCC) plus log of the total energy. We then compute and add first order and second order differences (deltas) for a total of 37 dimensions. We use 25 milisecond windows, sampled every 10 miliseconds.[1]

### 4.1.2 Synthetically spoken COCO

We generated synthetic speech for the captions in the MS COCO dataset (Lin et al., 2014) via the Google Text-to-Speech API.[2] The audio and the corresponding MFCC features are released as Chrupała et al. (2017)[3]. This TTS system we used produces high-quality realistic-sounding speech. It is nevertheless much simpler than real human speech as it uses a single voice, and lacks tempo variation or ambient noise. The data consists of over 300,000 images, each with five spoken captions. Five thousand images each are held out for validation and test. We use the splits and image features provided by Vendrov et al. (2015).[4] The image features also come from the VGG-16 network, but are averages of feature vectors for ten crops of each image. For the MS COCO captions we extracted only plain MFCC and total energy features, and did not add deltas in order to keep the amount of computation manageable given the size of the dataset.

## 4.2 Image retrieval

We evaluate our model on the task of ranking images given a spoken utterance, such that highly ranked images contain scenes described by the utterance. The performance on this task on validation data is also used to choose the best variant of the model architecture and to tune the hyperparameters. We compare the speech models to models trained on written sentences split into words. The best settings found for the four models were the following:

**Flickr8K Text RHN** 300-dimensional word embeddings, 1 hidden layer with 1024 dimensions, 1 microstep, initial learning rate 0.001.

**Flick8K Speech RHN** convolutional layer with length 6, size 64, stride 2, 4 hidden layers with 1024 dimensions, 2 microsteps, attention MLP with 128 hidden units, initial learning rate 0.0002

**COCO Text RHN** 300-dimensional word embeddings, 1 hidden layer with 1024 dimensions, 1 microstep, initial learning rate 0.001

**COCO Speech RHN** convolutional layer with length 6, size 64, stride 3, 5 hidden layers with 512 dimensions, 2 microsteps, attention MLP with 512 hidden units, initial learning rate 0.0002

All models were optimized with Adam (Kingma and Ba, 2014) with early stopping: we kept the parameters for the epoch which showed the best recall@10 on validation data.

| Model | R@1 | R@5 | R@10 | $\tilde{r}$ |
|---|---|---|---|---|
| Speech RHN$_{4,2}$ | 0.055 | 0.163 | 0.253 | 48 |
| Spectr. CNN | - | - | 0.179 | - |
| Text RHN$_{1,1}$ | 0.127 | 0.364 | 0.494 | 11 |

Table 1: Image retrieval performance on Flickr8K. R@N stands for recall at N; $\tilde{r}$ stands for median rank of the correct image.

| Model | R@1 | R@5 | R@10 | $\tilde{r}$ |
|---|---|---|---|---|
| Speech RHN$_{5,2}$ | 0.111 | 0.310 | 0.444 | 13 |
| Text RHN$_{1,1}$ | 0.169 | 0.421 | 0.565 | 8 |

Table 2: Image retrieval performance on MS COCO. R@N stands for recall at N; $\tilde{r}$ stands for median rank of the correct image.

Table 1 shows the results for the human speech from the Flickr8K dataset. The Speech RHN model scores substantially higher than model of Harwath and Glass (2015) on the same data. However the large gap between its perfomance and the scores of the text model suggests that Flickr8K is rather small for the speech task. In Table 2 we present the results on the dataset of synthetic speech from MS COCO. Here the text model is still better, but the gap is much smaller than for Flickr8K. We attribute this to the much larger size of dataset, and to the less noisy and less variable synthetic speech.

While the MS COCO text model is overall better than the speech model, there are cases where it outperforms the text model. We listed the top hundred cases where the ratio of the ranks of the correct image according to the two models was the smallest, as well as another hundred cases where it was the largest. Manual inspection did not turn

---

[1] We noticed that for a number of utterances the audio signal was very long: on inspection it turned out that most of these involved failure to switch off the microphone on the part of the workers, and the audio contained ambient noise or unrelated speech. We thus trucated all audio for this dataset at 10,000 miliseconds.

[2] Available at https://github.com/pndurette/gTTS.

[3] Available at https://doi.org/10.5281/zenodo.400926.

[4] See https://github.com/ivendrov/order-embedding.

up any obvious patterns for the cases of text being better than speech. For the cases where speech outperformed text, two patterns stood out: (i) sentences with spelling mistakes, (ii) unusually long sentences. For example for the sentence *a yellow*

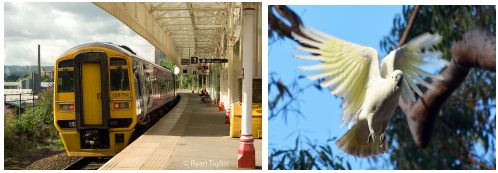

Figure 1: Images returned for utterance *a yellow and white birtd is in flight* by the text (left) and speech (right) models.

*and white birtd is in flight* the text model misses the misspelled word *birtd* and returns an irrelevant image, while the speech model seems robust to some degree of variation in pronunciation and returns the target image at rank 1 (see Figure 1). In an attempt to quantify this effect we counted the number of unique words with training set frequencies below 5 in the top 100 utterances with lowest and highest rank ratio: for the utterances where text was better there were 16 such words; for utterances where speech was better there were 28, among them misspellings such as *streeet*, *scears* (for *skiers*), *contryside*, *scull*, *birtd*, *devise*.

The distribution of utterance lengths in Figure 2 confirms pattern (ii): the set of 100 sentences where speech beats text by a large margin are longer on average and there are extremely long outliers among them. One of them is the 36-word-long utterance depicted in Figure 3, with ranks 470 and 2 for text and speech respectively. We suspect that the speech model's attention mechanism enables it to cherry pick key fragments of such monster utterances, while the text model lacking this mechanism may struggle. Figure 3 shows the plot of the attention weights for this utterance from the speech model.

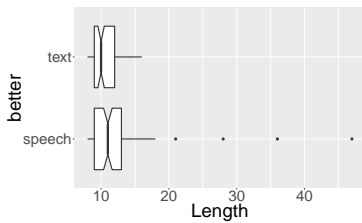

Figure 2: Length distribution for sentences where one model performs much better than the other.

### 4.3 Predicting utterance length

Our first auxiliary task is to predict the length of the utterance, using the features explained at the beginning of Section 4. Since the length of an utterance directly corresponds to how long it takes to articulate, we also use the number of time steps[5] as a feature and expect it to provide the upper bound for our task, especially for synthetic speech. We use a Ridge Regression model for predicting utterance length using each set of features. The model is trained on 80% of the sentences in the validation set, and tested on the remaining 20%. For all features regularization penalty $\alpha = 1.0$ gave the best results.

Figure 4 shows the results for this task on human speech from Flickr8K and synthetic speech from COCO. With the exception of the average input vectors for Flickr8K, all features can explain a high proportion of variance in the predicted utterance length. The pattern observed for the two datasets is slightly different: due to the systematic conversion of words to synthetic speech in COCO, using the number of time steps for this dataset yields the highest $R^2$. However, this feature is not as informative for predicting the utterance length in Flickr8K due to noise and variation in human speech, and is in fact outperformed by some of the features extracted from the model. Also, the input vectors from COCO are much more informative than Flickr8K due to larger quantity and simpler structure of the speech signal. However, in both datasets the best (non-ceiling) performance is obtained by using average unit activations from the hidden layers (layer 2 for COCO, and layers 3 and 4 for Flickr8K). These features outperform utterance embeddings, which are optimized according to the visual grounding objective of the model and most probably learn to ignore the superficial characteristics of the utterance that do not contribute to matching the corresponding image.

Note that the performance on COCO plateaus after the second layer, which might suggest that form-based knowledge is learned by lower layers. Since Flickr8K is much smaller in size, the stabilising happens later in layer 3.

---

[5]This is approximately $\frac{\text{duration in milliseconds}}{10 \times \text{stride}}$.

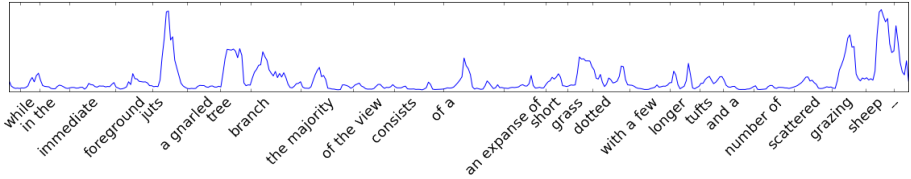

Figure 3: Attention weight distribution for a long utterance.

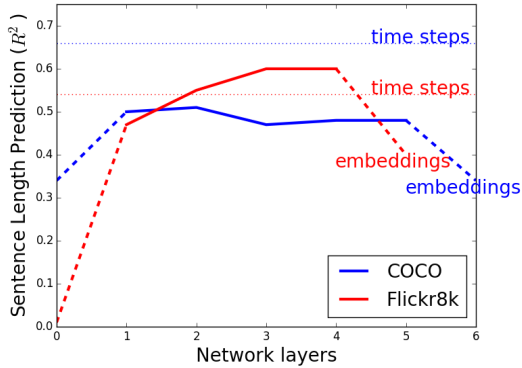

Figure 4: $R^2$ values for predicting utterance length for Flickr8K and COCO. Layers 1–5 represent (normalized) average unit activation, whereas the first (#0) and last point represent average input vectors and utterance embeddings, respectively.

### 4.4 Predicting word presence

Results from the previous experiment suggest that our model acquires information about higher level building blocks (words) in the continuous speech signal. Here we explore whether it can detect the presence or absence of individual words in an utterance. We formulate detecting a word in an utterance as a binary classification task, for which we use a multi-layer perceptron with a single hidden layer of size 1024, optimized by Adam. The input to the model is a concatenation of the feature vector representing an utterance and the one representing a target word. We again use utterance embeddings, average unit activations on each layer, and average input vectors as features, and represent each target word as a vector of MFCC features extracted from the audio signal synthetically produced for that word.

For each utterance in the validation set, we randomly pick one positive and one negative target (i.e., one word that does and one that does not appear in the utterance) that is not a stop word. To balance the probability of a word being positive or negative, we use each positive target as a negative target for another utterance in the validation set. The MLP model is trained on the positive and negative examples corresponding to 80% of the utterances in the validation set of each dataset, and evaluated on the remaining 20%.

Figure 5 shows the mean accuracy of the MLP on Flickr8K and COCO. All results using features extracted from the model are above chance (0.5), with the average unit activations of the hidden layers yielding the best results (0.65 for Flickr8K on layer 3, and 0.79 for COCO on layer 4). These numbers show that the speech model infers reliable information about word-level blocks from the low-level audio features it receives as input. The observed trend is similar to the previous task: average unit activations on the higher-level hidden layers are more informative for this task than the utterance embeddings, but the performance plateaus before the topmost layer.

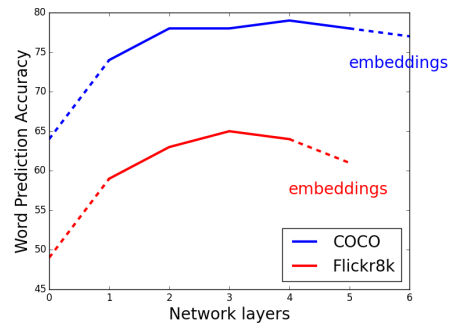

Figure 5: Mean accuracy values for predicting the presence of a word in an utterance for Flickr8K and COCO. Layers 1–5 represent the (normalized) average unit activations, whereas the first (#0) and last point represent average input vectors and utterance embeddings, respectively.

### 4.5 Sentence similarity

Next we explore to what extent the model's representations correspond to those of humans. We employ the Sentences Involving Compositional Knowledge (SICK) dataset (Marelli et al., 2014). SICK consists of image descriptions taken from Flickr8K and video captions from the SemEval 2012 STS MSRVideo Description data set (STS) (Agirre et al., 2012). Captions were paired at random, as well as modified to obtain semantically similar and contrasting counterparts, and the resulting pairs were rated for semantic similarity.

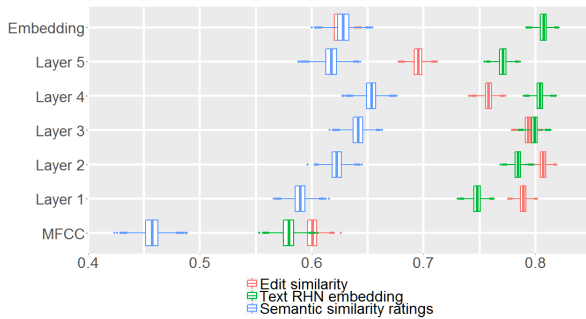

Figure 6: Pearson's $r$ of cosine similarities of averaged input MFCCs and COCO Speech RHN hidden layer activation vectors and embeddings of sentence pairs with relatedness scores from SICK, cosine similarity of COCO Text RHN embeddings, and edit similarity.

For all sentence pairs in SICK, we generate synthetic spoken sentences and feed them to the COCO Speech RHN, and calculate the cosine similarity between the averaged MFCC input vectors, the averaged hidden layer activation vectors, and the sentence embeddings. Z-score transformation was applied before calculating the cosine similarities. We then correlate these cosine similarities with

- semantic relatedness according to human ratings
- cosine similarities according to z-score transformed embeddings from COCO Text RHN
- *edit similarities*, a measure of how similar the sentences are in form, specifically, $1-$normalized Levenshtein distance over character sequences

Figure 6 shows a boxplot over 10,000 bootstrap samples for all correlations. We observe that (i) correlation with edit similarity initially increases, then decreases; (ii) correlation with human relatedness scores and text model embeddings increases until layer 4, but decreases for hidden layer 5. The initially increasing and then decreasing correlation with edit similarity is consistent with the findings that information about form is encoded by lower layers. The overall growing correlation with both human semantic similarity ratings and the COCO Text RHN indicate that higher layers learn to represent semantic knowledge. We were somewhat surprised by the pattern for the correlation with human ratings and the Text model similarities which drops for layer 5. We suspect it may be caused by the model at this point in the layer hierarchy being strongly tuned to the specifics of the COCO dataset. To test this, we checked the correlations with COCO Text embeddings on validation sentences from the COCO dataset instead of SICK. These increased monotonically, in support of our conjecture.

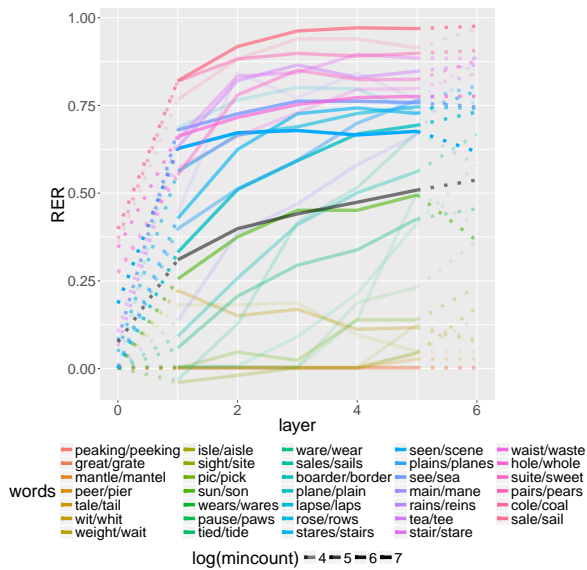

Figure 7: Disambiguation performance per layer. Points #0 and #6 (connected via dotted lines) represent the input vectors and utterance embeddings, respectively. The black line shows the overall mean RER.

### 4.6 Homonym disambiguation

Next we simulate the task of distinguishing between pairs of homonyms, i.e. words with the same acoustic form but different meaning. We group the words in the union of the training and validation data of the COCO dataset by their phonetic transcription. We then pick pairs of words which have the same pronunciation but different spelling, for example *suite/sweet*. We impose the following conditions: (a) both forms appear more than 20 times, (b) the two forms have different meaning (i.e. they are not simply variant spellings like *theater/theatre*), (c) neither form is a function word, and (d) the more frequent form constitutes less than 95% of the occurrences. This gives us 34 word pairs. For each pair we generate a binary classification task by taking all the utterances where either form appears, using average input vectors, utterance embeddings, and average unit activations as features. Instances for all feature sets are normalized to unit L2 norm.

For each task and feature set we run stratified 10-fold cross validation using Logistic Regression to predict which of the two words the utterance contains. Figure 7 shows, for each pair, the relative error reduction of each feature set with respect to the majority baseline. There is substantial variation across word pairs, but overall the task becomes easier as the features come from higher layers in the network. Some forms can be disambiguated with very high accuracy (e.g. *sale/sail, cole/coal, pairs/pears*), while some others cannot be distinguished at all (*peaking/peeking, great/grate, mantle/mantel*). We examined the sentences containing the failing forms, and found out that almost all occurrences of *peaking* and *mantle* were misspellings of *peeking* and *mantel*, which explains the impossibility of disambiguating these cases.

## 5 Conclusion

We present a multi-layer recurrent highway network model of language acquisition from visually grounded speech signal. Through detailed analysis we uncover how information in the input signal is transformed as it flows through the network: formal aspects of language such as word identities that not directly present in the input are discovered and encoded low in the layer hierarchy, while semantic information is most strongly expressed in the topmost layers.

Going forward we would like to compare the representations learned by our model to the brain activity of people listening to speech in order to determine to what extent the patterns we found correspond to localized processing in the human cortex. This will hopefully lead to a better understanding of language learning and processing by both artificial and neural networks.

## Acknowledgements

We would like to thank David Harwath for making the Flickr8k Audio Caption Corpus publicly available.

## References

Yossi Adi, Einat Kermany, Yonatan Belinkov, Ofer Lavi, and Yoav Goldberg. 2016. Fine-grained analysis of sentence embeddings using auxiliary prediction tasks. *arXiv preprint arXiv:1608.04207* .

Eneko Agirre, Mona Diab, Daniel Cer, and Aitor Gonzalez-Agirre. 2012. Semeval-2012 task 6: A pilot on semantic textual similarity. In *Proceedings of the First Joint Conference on Lexical and Computational Semantics*. Association for Computational Linguistics, volume 2, pages 385–393.

Raffaella Bernardi, Ruket Cakici, Desmond Elliott, Aykut Erdem, Erkut Erdem, Nazli Ikizler-Cinbis, Frank Keller, Adrian Muscat, and Barbara Plank. 2016. Automatic description generation from images: A survey of models, datasets, and evaluation measures. *arXiv preprint arXiv:1601.03896* .

Grzegorz Chrupała, Akos Kádár, and Afra Alishahi. 2015. Learning language through pictures. In *Proceedings of the 53rd Annual Meeting of the Association for Computational Linguistics*.

Grzegorz Chrupała, Lieke Gelderloos, and Afra Alishahi. 2017. Synthetically spoken COCO. https://doi.org/10.5281/zenodo.400926.

Afsaneh Fazly, Afra Alishahi, and Suzanne Stevenson. 2010. A probabilistic computational model of cross-situational word learning. *Cognitive Science: A Multidisciplinary Journal* 34(6):1017–1063.

Michael C. Frank, Noah D. Goodman, and Joshua B. Tenenbaum. 2007. A Bayesian framework for cross-situational word-learning. In *Advances in Neural Information Processing Systems*. volume 20.

Lieke Gelderloos and Grzegorz Chrupała. 2016. From phonemes to images: levels of representation in a recurrent neural model of visually-grounded language learning. In *Proceedings of COLING 2016, the 26th International Conference on Computational Linguistics: Technical Papers*.

David Harwath and James Glass. 2015. Deep multimodal semantic embeddings for speech and images. In *IEEE Automatic Speech Recognition and Understanding Workshop*.

David Harwath and James R Glass. 2017. Learning word-like units from joint audio-visual analysis. *arXiv preprint arXiv:1701.07481* .

David Harwath, Antonio Torralba, and James Glass. 2016. Unsupervised learning of spoken language with visual context. In *Advances in Neural Information Processing Systems*. pages 1858–1866.

Kaiming He, Xiangyu Zhang, Shaoqing Ren, and Jian Sun. 2015. Deep residual learning for image recognition. *arXiv:1512.03385* .

Micah Hodosh, Peter Young, and Julia Hockenmaier. 2013. Framing image description as a ranking task: Data, models and evaluation metrics. *Journal of Artificial Intelligence Research* 47:853–899.

Ákos Kádár, Grzegorz Chrupała, and Afra Alishahi. 2016. Representation of linguistic form and function in recurrent neural networks. *CoRR* abs/1602.08952.

Andrej Karpathy and Li Fei-Fei. 2015. Deep visual-semantic alignments for generating image descriptions. In *Proceedings of the IEEE Conference on Computer Vision and Pattern Recognition*. pages 3128–3137.

Diederik P. Kingma and Jimmy Ba. 2014. Adam: A method for stochastic optimization. *CoRR* abs/1412.6980.

Angeliki Lazaridou, Grzegorz Chrupała, Raquel Fernández, and Marco Baroni. 2016. Multimodal semantic learning from child-directed input. In *The 15th Annual Conference of the North American Chapter of the Association for Computational Linguistics: Human Language Technologies*.

Jiwei Li, Will Monroe, and Dan Jurafsky. 2016. Understanding neural networks through representation erasure. *arXiv preprint arXiv:1612.08220* .

Tsung-Yi Lin, Michael Maire, Serge Belongie, James Hays, Pietro Perona, Deva Ramanan, Piotr Dollár, and C Lawrence Zitnick. 2014. Microsoft coco: Common objects in context. In *Computer Vision–ECCV 2014*, Springer, pages 740–755.

Marco Marelli, Stefano Menini, Marco Baroni, Luisa Bentivogli, Raffaella Bernardi, and Roberto Zamparelli. 2014. A sick cure for the evaluation of compositional distributional semantic models. In *LREC*. pages 216–223.

Yajie Miao, Jinyu Li, Yongqiang Wang, Shi-Xiong Zhang, and Yifan Gong. 2016. Simplifying long short-term memory acoustic models for fast training and decoding. In *IEEE International Conference on Acoustics, Speech and Signal Processing (ICASSP)*. IEEE, pages 2284–2288.

Aaron van den Oord, Nal Kalchbrenner, and Koray Kavukcuoglu. 2016. Pixel recurrent neural networks. *arXiv preprint arXiv:1601.06759* .

Deb K Roy and Alex P Pentland. 2002. Learning words from sights and sounds: a computational model. *Cognitive Science* 26(1):113 – 146.

Olga Russakovsky, Jia Deng, Hao Su, Jonathan Krause, Sanjeev Satheesh, Sean Ma, Zhiheng Huang, Andrej Karpathy, Aditya Khosla, Michael Bernstein, Alexander C. Berg, and Li Fei-Fei. 2014. ImageNet Large Scale Visual Recognition Challenge.

Karen Simonyan and Andrew Zisserman. 2014. Very deep convolutional networks for large-scale image recognition. *CoRR* abs/1409.1556.

Jeffrey M. Siskind. 1996. A computational study of cross-situational techniques for learning word-to-meaning mappings. *Cognition* 61(1-2):39–91.

Gabriel Synnaeve, Maarten Versteegh, and Emmanuel Dupoux. 2014. Learning words from images and speech. In *NIPS Workshop on Learning Semantics, Montreal, Canada*.

Zhiyuan Tang, Ying Shi, Dong Wang, Yang Feng, and Shiyue Zhang. 2016. Memory visualization for gated recurrent neural networks in speech recognition. *arXiv preprint arXiv:1609.08789* .

Ivan Vendrov, Ryan Kiros, Sanja Fidler, and Raquel Urtasun. 2015. Order-embeddings of images and language. *arXiv preprint arXiv:1511.06361* .

Chen Yu and Dana H Ballard. 2004. A multimodal learning interface for grounding spoken language in sensory perceptions. *ACM Transactions on Applied Perception (TAP)* 1(1):57–80.

Bolei Zhou, Agata Lapedriza, Jianxiong Xiao, Antonio Torralba, and Aude Oliva. 2014. Learning deep features for scene recognition using places database. In Z. Ghahramani, M. Welling, C. Cortes, N. D. Lawrence, and K. Q. Weinberger, editors, *Advances in Neural Information Processing Systems 27*, Curran Associates, Inc., pages 487–495.

Julian Georg Zilly, Rupesh Kumar Srivastava, Jan Koutník, and Jürgen Schmidhuber. 2016. Recurrent highway networks. *arXiv preprint arXiv:1607.03474* .